\documentclass[review]{elsarticle}

\usepackage{lineno,hyperref}
\modulolinenumbers[5]

\journal{Pattern Recognition}

\usepackage{graphicx}
\usepackage{multirow}
\usepackage{epstopdf}
\usepackage{amsmath}
\usepackage{amssymb}
\usepackage{amsfonts}
\usepackage{color}
\usepackage{subfigure}
\usepackage{verbatim}
\usepackage[ruled,linesnumbered]{algorithm2e}
\usepackage{adjustbox,lipsum}
\usepackage{dsfont}
\usepackage{booktabs}
\usepackage{multirow}
\usepackage{array}

\usepackage{amssymb}

\usepackage[figuresright]{rotating}

\newcommand{\etal}{{\em et al.\,}}       
\newcommand{\eg}{{\em e.g.}}           
\newcommand{\ie}{{\em i.e.}}           
\newcommand{\rev}[1]{\textcolor{black}{#1}}

\begin{document}

\begin{frontmatter}




\title{Better Pseudo-label: Joint Domain-aware Label and Dual-classifier for Semi-supervised Domain Generalization}


\author[nju,ins]{Ruiqi Wang\corref{cor3}}
\ead{wangrq@smail.nju.edu.cn}
\author[seu]{Lei Qi\corref{cor3}}
\ead{qilei@seu.edu.cn}
\author[nju,ins]{Yinghuan Shi\corref{cor1}}
\ead{syh@nju.edu.cn}
\author[nju,ins]{Yang Gao}
\ead{gaoy@nju.edu.cn}

\cortext[cor1]{Corresponding author: Yinghuan Shi.}
\cortext[cor3]{Co-first authors: Ruiqi Wang and Lei Qi.}

\address[nju]{State Key Laboratory for Novel Software Technology, Nanjing University, Nanjing, China}
\address[ins]{National Institute of Healthcare Data Science, Nanjing University, Nanjing, China}
\address[seu]{School of Computer Science and Engineering, Key Lab of Computer Network and Information Integration (Ministry of Education), Southeast University, Nanjing, China}

\begin{abstract}
With the goal of directly generalizing trained model to unseen target domains, domain generalization (DG), a newly proposed learning paradigm, has attracted considerable attention. Previous DG models usually require a sufficient quantity of annotated samples from observed source domains during training. In this paper, we relax this requirement about full annotation and investigate semi-supervised domain generalization (SSDG) where only one source domain is fully annotated along with the other domains totally unlabeled in the training process. With the challenges of tackling the domain gap between observed source domains and predicting unseen target domains, we propose a novel deep framework via joint domain-aware labels and dual-classifier to produce high-quality pseudo-labels. Concretely, to predict accurate pseudo-labels under domain shift, a domain-aware pseudo-labeling module is developed. Also, considering inconsistent goals between generalization and pseudo-labeling: former prevents overfitting on all source domains while latter might overfit the unlabeled source domains for high accuracy, we employ a dual-classifier to independently perform pseudo-labeling and domain generalization in the training process. 
When accurate pseudo-labels are generated for unlabeled source domains, the domain mixup operation is applied to augment new domains between labeled and unlabeled domains, which is beneficial for boosting the generalization capability of the model. 
Extensive results on publicly available DG benchmark datasets show the efficacy of our proposed SSDG method.
\end{abstract}

\begin{keyword}
Semi-supervised learning \sep Domain generalization \sep Image recognition\sep Feature representation
\end{keyword}

\end{frontmatter}
\section{Introduction}
\label{sec:introduction}

Nowadays, with the development of data acquisition, current data are frequently captured from multiple sources (\eg, video, image, text), generated from various contributors (\eg, different artists), or collected from multiple sites (\eg, different data centers), making the distribution shift between different modalities or sites usually occurs \cite{seguda, robustreid}. Therefore, due to the distribution shift, the model trained on training data or source domains could perform poorly on test data or target domains. To address this limitation, a new setting namely domain generalization (DG), aiming to train model on observed source domains for directly generalizing to arbitrary unseen target domains, is becoming a hot topic with increasing interests.

According to our investigation, unfortunately, most current DG models belong to supervised setting where multiple fully labeled source domains are the prerequisite before training DG models. 
As we known, high-quality labels are often expensive and laborious to obtain, which drives us to alleviate the label requirement in the observed source domains.

\begin{figure}[t]
\begin{center}
\includegraphics[width=0.8\linewidth]{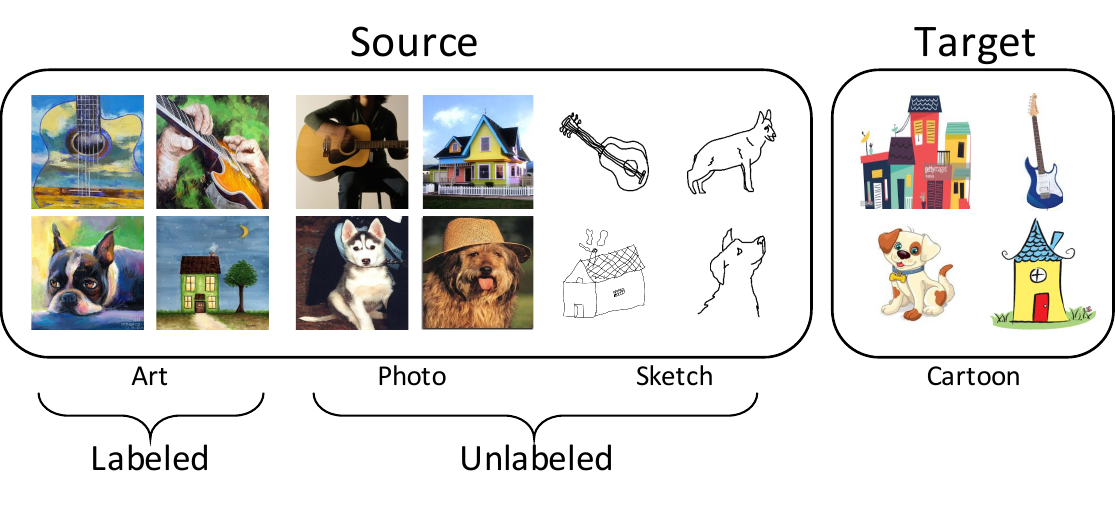}
\end{center}
   \caption{Unlike typical conventional domain generalization, semi-supervised domain generalization takes both the labeled and unlabeled source domains as input, aiming to train an adaptive model for the unseen target domain.}
\label{fig:semiDG}
\end{figure}

Formally, we here name our setting---first training the model with both labeled source domains and unlabeled source domains and then performing prediction on unseen target domains---as semi-supervised domain generalization (SSDG in short). This setting owns its practical meaning. For example, in real-world applications, there are a large number of totally unlabeled datasets (\ie, web-crawled datasets, massive data in data center). The advantage of our setting is that arbitrary unlabeled domains can be utilized to cooperate with the labeled domains for benefiting domain generalization in a free lunch way. We show this setting in Figure \ref{fig:semiDG}. Particularly, in this paper we merely consider the case that only one source domain is fully labeled (along with several unlabeled source domains) in the training stage. The `one labeled source domain' case is more practical because annotating samples is difficult, expensive and time-consuming. And a huge amount of unlabeled data can be easily obtained in real-world applications. However, the `one labeled source domain' case is more challenging because we need to generate pseudo-labels for extensive samples 
from multiple unlabeled domains with different data distributions.

Since unlabeled samples in source domains are abundant and each unlabeled sample actually belongs to a specific yet unknown class, we consider assigning pseudo-labels to unlabeled samples. Intuitively, the accuracy of pseudo-labels largely affects the final results.
The pseudo-labeling technique \cite{lee2013pseudo} has shown its effectiveness in conventional semi-supervised learning (SSL) problems by iteratively using higher confident samples to aid subsequent learning on lower confident samples. However, compared with conventional SSL, producing high-quality pseudo-labels in SSDG is much more challenging due to the following two reasons:
\begin{enumerate}
    \item The domain shift between observed labeled and unlabeled source domains is definitely a negative factor for accurate pseudo-labels, which may lead to a drastic performance degeneration. 
    \item Since there is the unpredictable domain discrepancy between unlabeled source domains and the unseen target domain, a generalizable model which well fits the target domain may suffer a drop of accuracy on unlabeled source domains.
\end{enumerate}
Considering these two issues, we develop two improvements for accurate pseudo-label prediction.

\begin{figure}[t]
\centering
\includegraphics[width=0.95\columnwidth]{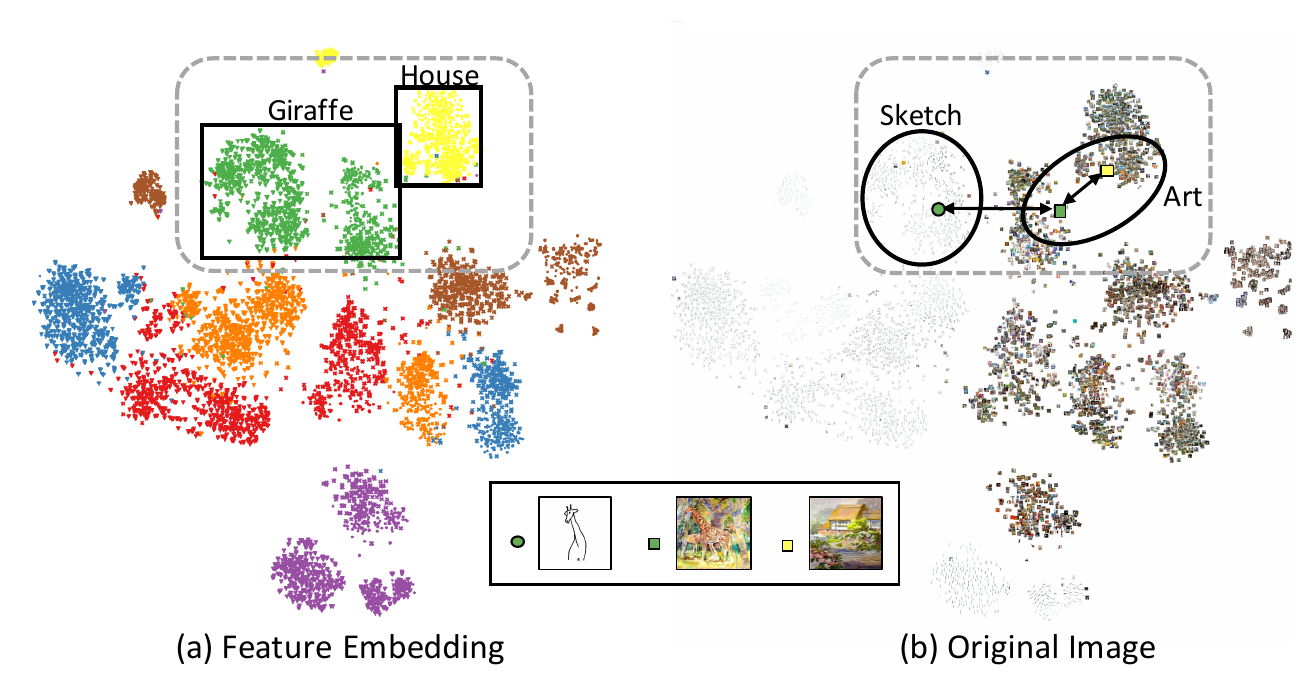}
\caption{(a): Visualization of feature embeddings from a fully supervised DG model using a domain discriminator to reduce the domain gap on PACS. Note that different colors denote different classes. (b): The original images from PACS, which correspond to features in (a) from the position view. We focus on giraffe and house, Sketch domain, and Art painting domain. As seen, after feature alignment, features are still not strictly domain-invariant, \eg, several giraffe samples from the Art are closer to some house samples from the same domain than a giraffe from the Sketch.}
\label{fig:vi}
\end{figure}

Firstly, we propose \textit{domain-aware pseudo-labeling method} to improve the quality of pseudo-labels under domain shift. 
As aforementioned, the domain shift between labeled and unlabeled source domains deteriorates the accuracy of pseudo-labels.
In Figure \ref{fig:vi}, we visualize the feature distribution of the DANN \cite{grl} model trained via fully supervised learning on PACS. As observed, the experimental result shows that samples have been well mapped to their categories, whereas inside a typical class, features from different domains intra a class are separated. Therefore, to obtain more accurate pseudo-labels, we iteratively maintain the average feature of the most confident unlabeled samples for each class of each domain in the memory, which is used as \textit{domain-aware class representation}. Afterwards, when assigning pseudo-labels to unlabeled samples, we combine the output probability of the classifier with the similarity to its class representation to decide which class it belongs to.

Secondly, in SSDG, our goal is to improve the generalization ability of the network to adapt to an arbitrary target domain. However, intuitively, a generalizable model could underfit the unlabeled training domains, thus the accuracy of pseudo-labels could decrease.
Considering inconsistent goals between generalization and pseudo-labeling---former prevents to overfit source domains while latter might overfit specific domain for high accuracy, we propose to use a dual-classifier to avoid the possible accuracy degradation of pseudo-labels, which leverages the independent classifiers for joint pseudo-label assignment and domain generalization. In our dual-classifier network, the two branches are trained with different objective functions but a shared feature extractor.

Our contributions can be summarized as follows: 
\begin{itemize}
    \item We propose an effective framework that can be trained in an end-to-end manner to obtain more accurate pseudo-labels of unlabeled data for the semi-supervised domain generalization task. 
    \item We develop the domain-aware pseudo-labeling module to handle the domain shift during generating pseudo-labels. Also, the dual-classifier is proposed to mitigate the conflict between the DG task and the pseudo-label generation. 
    \item Extensive experiments on benchmark datasets, \ie, PACS, OfficeHome, miniDomainNet and VLCS, show the effectiveness of our method compared with several baselines and the state-of-the-art methods.
\end{itemize}

\section{Related Works}\label{s-related}
We review the recent work about unsupervised domain adaptation, domain generalization, semi-supervised learning, and semi-supervised domain generalization.
\subsection{Unsupervised Domain Adaptation}
Unsupervised domain adaptation can effectively transfer knowledge from an annotated source domain to an unlabeled target domain. Mainstream approaches include discrepancy-based \cite{DAN_,reda,swd,mutual,mdd, bures}, adversarial-based \cite{DANN,ADDA,CAN,hybrid_adversarial_network,aaa} and pseudo-labeling-based \cite{Chen2019ProgressiveFA,zhang2021adversarial} methods. 
Lee \etal \cite{swd} design sliced Wasserstein discrepancy (SWD) to capture the discrepancy between the outputs of task-specific classifiers. Li \etal \cite{mdd} propose maximum density divergence (MDD) to measure the distribution divergence and apply MDD to minimize the inter-domain discrepancy and maximize the intra-class density. Zhang \etal \cite{hybrid_adversarial_network} introduce a novel Hybrid Adversarial Network (HAN), which achieves a joint adversarial learning with class information and domain alignment. Zhang \etal \cite{zhang2021adversarial} propose to increase the robustness of the model by incorporating high-confidence samples from the target domain. 
Li \etal \cite{aaa} propose a more practical UDA setting where either the source data or the target data are unknown, and handle the UDA setting by the adversarial attack.
In addition, some recent methods aim to address the issue of limited computing power in UDA problems. Li \etal \cite{fan} propose the Faster Domain Adaptation (FDA) protocol to accelerate unsupervised domain adaptation. Despite UDA being related to DG, UDA has access to the target domain while DG cannot observe the target domain during training.

\subsection{Domain Generalization}
Domain generalization methods can be substantially categorized into data-based methods, feature-based methods, and learning strategy-based methods \cite{DGsurvey}. The data-based methods aim to generate virtual training data for a more generalizable model, \eg, the methods in \cite{databaseone, augtwo, augthree, augone, tracking}
enlarge the training set by image generation and data augmentation techniques which are applied to solve data insufficiency\cite{covid}. 
Feature-based solutions \cite{featurebaseone, featurebasetwo, invariantone, invariancetwo} extract domain-agnostic representations on multi-source domains. 
Another promising technique for domain generalization is meta-learning, such as \cite{metaone, metatwo, segDG}. Besides, some methods based on other learning strategies ({\em e.g.}, self-supervision, Ensemble learning) are proposed to obtain the generalizable model, including \cite{ensemble_learning, self_supervison}.

\subsection{Semi-supervised Learning}
Current semi-supervised learning methods can be roughly classified into three categories, {\em i.e.}, entropy regularization based methods, pseudo-label based methods, and consistency regularization based methods. The essence of all these three categories is to force a low-density distribution between different classes \cite{2019S4L}. A straightforward way is to add a loss term to directly and explicitly reduce the entropy of the predictions on unlabeled data. Entropy regularization \cite{entropyMinimization} encourages a confident prediction on unlabeled data by minimizing the entropy of the predictions of unlabeled data.
Pseudo-label based methods \cite{lee2013pseudo, selflabel, inconsistencyaware} assign approximate classes to unlabeled samples by the inference of the model trained on labeled samples.
Consistency regularization shows great success more recently, which includes $\pi$-Model \cite{temporalensembling}, Temporal Ensembling \cite{temporalensembling} and Mean Teacher \cite{meanteacher}, etc.
Besides, a series of holistic approaches to semi-supervised learning have obtained state-of-the-art performance on commonly-studied SSL benchmarks recently. Unsupervised Data Augmentation (UDA) \cite{uda} 
improves consistency loss by substituting simple noising operations with advanced data augmentation, such as RandAugment \cite{randaug}. MixMatch \cite{mixmatch} unifies the existing data augmentation, pseudo-labeling, and mixup to achieve both consistency regularization and entropy regularization. ReMixMatch \cite{remixmatch} further improves MixMatch \cite{mixmatch}. FeatMatch \cite{featmatch} applies learned feature-based augmentation to consistency loss. FixMatch \cite{fixmatch} inherits UDA and ReMixMatch, combines pseudo-labeling and consistency regularization, and finally obtains good performance on SSL benchmarks. Differently, in our SSDG, there is a data-distribution discrepancy between labeled and unlabeled training data, thus these typical SSL methods could not effectively handle the issue.
\subsection{Semi-supervised Domain Generalization}
To the best of our knowledge, only very a few works have been proposed for semi-supervised domain generalization problem. 
DGSML \cite{semidgone} and StyleMatch \cite{stylematch} tackle a new setting in domain generalization problem, where the labeled samples in each domain are not fully labeled. Although we both assign pseudo-labels to unlabeled data, we solve two different scenarios and the challenges we face are totally different. DSDGN \cite{semidgtwo} solves a semi-supervised domain generalization problem which is similar to us. It applies a Wasserstein generative adversarial
network with gradient penalty based adversarial training framework to align feature embedding, and simply adopts the original pseudo-labeling method for unlabeled data. However, this method does not consider the domain shift during pseudo-labeling.

\section{Method}\label{s-framework}
\begin{figure*}[ht]
  \centering
  \includegraphics[width=0.98\linewidth]{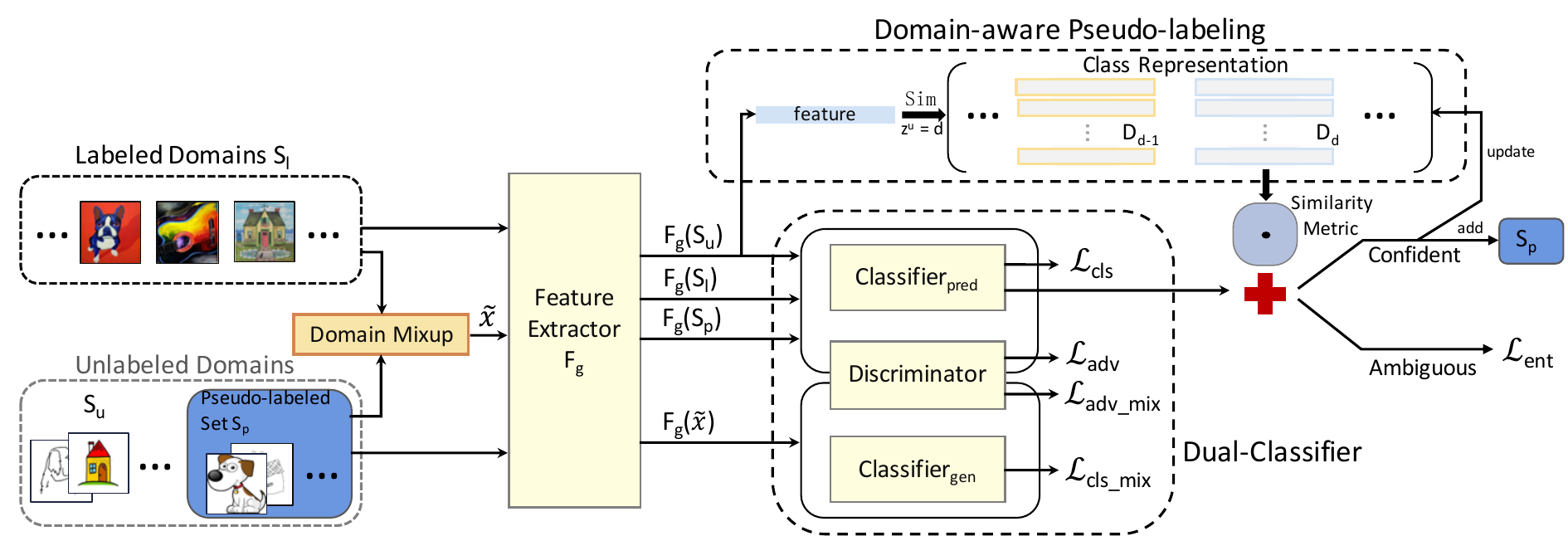}
  \caption{Illustration of our framework, which mainly consists of the feature extractor, domain-aware pseudo-labeling, dual-classifier and domain discriminator. During training, model optimization and pseudo-label prediction are alternate and iterative. 
}
  \label{framework}
\end{figure*}

Unlike the supervised domain generalization (DG) setting, as aforementioned, semi-supervised DG further alleviates the fully-labeled requirement and allows several source domains to be totally unlabeled during training. Formally, we now provide the notations used in our setting.
In SSDG, assume we have one labeled source domain in the labeled domain set $\mathcal{S}_{l} = \{D^l\}$, and $n$ unlabeled source domains in the unlabeled domain set $\mathcal{S}_{u}=\{D_1^u, ..., D_n^u\}$, and one target domain $D_t$. Note that, $D_t$ is not used in the training process. A training sample in the labeled source domain can be represented as a raw input $x$, a semantic label $y$, and a domain label $z$. Assuming that the number of the training samples in the labeled source domain is $n_l$, it can be denoted as $D^l=\{(x_i^l, y_i^l, z_i^l=0)\}_{i=1}^{n_l}$. And an unlabeled domain $D_j$ can be represented as $D_j=\{(x_i^u, z_i^u=j)\}_{i=1}^{n_j}$, when the number of the training samples in $D_j$ is $n_j$. $C$ stands for the number of categories in the classification dataset. The shared feature extractor, the predictive classifier, the generalizable classifier, and the domain classifier are denoted by $F_g$, $F_c$, $F_m$, and $F_d$, respectively. The overview of our method is illustrated in Figure \ref{framework}.

\subsection{Domain-aware Pseudo-labeling}\label{DAPL}
In the cross-domain scenario, a mixture of samples from all domains are fed into a shared classifier together. However, the data distributions of different domains are significantly different. Accordingly, for different domains, the discriminative characteristics that are critical to classification could be different. For instance, on the PACS dataset, images in \textit{photo} domain are color-specified, while images in \textit{cartoon} domain are not. In \textit{sketch} domain, the color information is totally erased, which makes samples in this domain even harder to distinguish. Thus, even we have applied a discriminator to align features from different domains, the features haven't been perfectly aligned yet. Due to the large and unpredictable domain gap between different domains, the classifier that well fits the labeled domain will generate the poor pseudo-label for unlabeled domains.

In order to alleviate the bias caused by the aforementioned domain gap, we propose domain-aware pseudo-labeling module in our framework. In particular, we first yield the \textit{domain-aware class representation} for each class of each unlabeled domain, which indicates the mean feature of the most highly confident samples for each class in each domain. Then, when generating the pseudo-label for each unlabeled sample by integrating 1) its predicted probability from the shared predictive classifier and 2) its largest similarity to the class representation from its domain, we can obtain the modified probability for the more reliable pseudo-label. 

For an unlabeled sample $(x^u, z^u)$, if the domain label $z^u$ is equal to $d$, it means that this sample is from the $d^{th}$ domain. Using $M^d \in \mathbb{R}^{D\times C}$
to denote the matrix by gathering the $D$-dimensional class representation from total $C$ classes in the $d^{th}$ domain. $\texttt{sim}(\cdot, \cdot)$ is a similarity measurement function. The similarity is conducted by calculating the cosine similarity between $F_g(x^u)$ and the class representation of each class, where $F_g(x^u)\in \mathbb{R}^{1\times D}$ is the D-dimensional feature embedding. $\psi(x^u) \in \mathbb{R}^{1\times C}$ is the $C$-dimension similarity vector obtained after a softmax function. Then we modify the predicted probability $q(x^u)$ by a correction term to form $s(x^u)$. This above process can be formulated as follows:

\begin{equation}
 s(x^u) = \gamma q(x^u) + (1-\gamma) \psi (x^u),
 \label{six}
\end{equation}
where $q(x^u) = F_c(F_g(x^u))$, $\psi (x^u) = \texttt{sim}(F_g(x^u), M^d)$.

Now the pseudo-label $\widehat{y^u}$ is assigned by $s(x^u)$ as same as the conventional pseudo-labeling way:
\begin{equation}
\widehat{y^u_i}=\left\{\begin{array}{ll}
1 & \text{if~} i=\mathrm{argmax}_{i^{\prime}} s_{i^{\prime}}(x^u) ~and~ s_{i^{\prime}}(x^u) > \delta \\
0 & \text {otherwise},
\end{array}\right.
\label{pseu}
\end{equation}
where $\delta$ is a threshold. If the final confidence of one sample is larger than the threshold, the pseudo-label is reliable, otherwise it is ambiguous.

The group of representation is updated at each epoch and used at the next epoch. We propose two policies for the domain-aware class representation production. The simplest way is to select the sample with the highest confidence predicted by the classifier as class representation at each epoch. However, the mislabeled sample could arise an accumulation of prediction error at the next epoch \cite{highconfidencewrong}. Furthermore, we propose to calculate an ensemble representation with some reliable samples of each class in each unlabeled domain. Specifically, we maintain a list of the highest confident samples for each class. We append one sample to the list only if its confidence is higher than the existing highest confidence in the list. And the capacity of each list is $k$, we stop appending samples if the size of the list reaches the limit. At the start of each epoch, 
we calculate the average of the samples (feature embeddings) in the list as class representation. In our experiments $k$ is set as 100. The comparison of the two schemes will be shown in our experiment.

 \begin{figure}[t]
\centering
\includegraphics[width=0.98\linewidth]{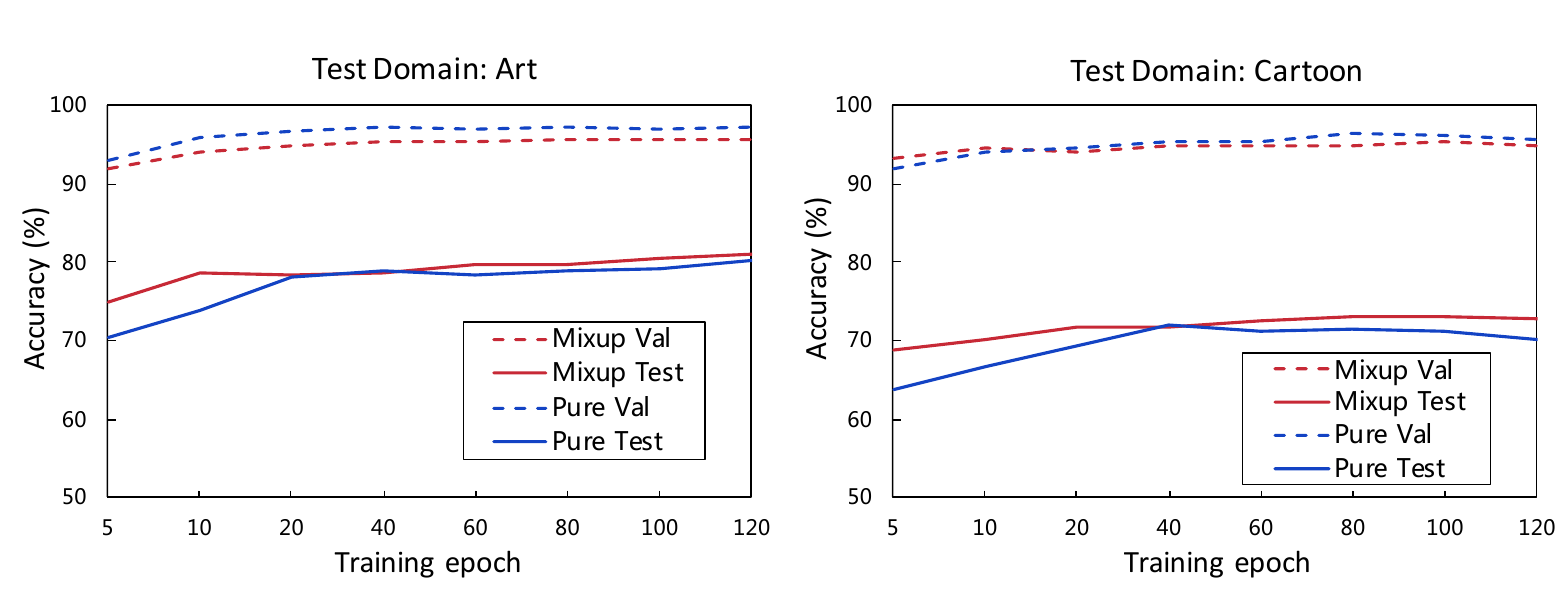}
\caption{Validation and test accuracy with mixup samples or with pure samples on PACS. Left: Train on Photo, Cartoon and Sketch and test on Art painting. Right: Train on Photo, Art painting and Sketch and test on Cartoon.}
\label{mixup-vs-nomixup}
\end{figure}

\subsection{Dual-classifier}\label{DC}
As known, overfitting empirically occurs when a model begins to fit the domain-specific characteristic in the training data rather than learning to generalize from a trend \cite{overfitting}. Consequently, the overfitting issue finally causes performance degradation on unseen test data. This empirical knowledge inspires us that a generalizable model which works well in domain generalization could be non-overfitting for the source training domains.

We train a model with domain mixup \cite{mixup} and without domain mixup in a supervised way and draw the comparison of both test and validation accuracy. As shown in Figure \ref{mixup-vs-nomixup}, by domain mixup, the model performs more accurately in the test domain, indicating that the model is more generalizable. However, the accuracy in the training domains drops, which is displayed by the validation accuracy. 

In our approach, we apply cross-domain mixup towards a more generalizable classifier. However, we observe that the generalizable DG classifier is not good at pseudo-labeling. Thus, we apply an auxiliary classifier to achieve the pseudo-labeling, and the classifier for giving the pseudo-labels is trained only by pure samples from source domains. Therefore, we utilize dual-classifier architecture to reduce the conflict between the DG task and the pseudo-label production.

\subsection{Mining the Knowledge of Unlabeled Domains}\label{mining}
When accurate pseudo-labels on unlabeled samples are generated, domain mixup is applied to confident samples with their pseudo-labels and the raw labeled data. For ambiguous samples, since pseudo-labels are not be assigned, the entropy loss is applied to make full use of these samples.

\textbf{Confident unlabeled samples.} As inspired by \cite{mixup}, we innovatively interpolate between a labeled domain and an unlabeled domain to achieve inter-domain data augmentation. Our intuition is that the inter-domain samples generated by domain mixup can boost the generalization of networks by introducing additional training domains, which has been verified as an important technique in domain generalization \cite{augone, augtwo, augthree}. 
Assuming that we have already assigned a pseudo-label for an unlabeled sample $(x^u, z^u)$, forming $(x^u, \widehat{y^u}, z^u)$.
And we have a sample from a labeled domain $(x^l, y^l, z^l)$. $\widehat{y^u}$, $z^u$, $y^l$, and $z^l$ are all one-hot vectors. Then the operation is formulated as below:
\begin{equation}
\tilde{x}=\lambda {x}^{l}+(1-\lambda){x}^{u},
\label{one}
\end{equation}
\begin{equation}
\tilde{y}=\lambda {y}^{l}+(1-\lambda)\widehat{y^u},
\label{two}
\end{equation}
where $\lambda \sim \mathrm{Beta}(\alpha, \alpha)$, for $\alpha > 0$, and $\lambda \in [0,1]$. The hyper-parameter $\alpha$ controls the strength of interpolation.

Additionally, to further enhance the generalization ability, we also apply mixup to domain labels of labeled and unlabeled samples for training domain discriminator as follows:
\begin{equation}
\tilde{z}=\lambda {z}^{l}+(1-\lambda){z}^{u},
\label{eq3}
\end{equation}
where $\lambda$ is same with that used in Eqn. (\ref{one}) and Eqn. (\ref{two}). According to these steps, we could finally generate a virtual sample $(\tilde{x}, \tilde{y}, \tilde{z})$.

\textbf{Ambiguous unlabeled samples.} For the unlabeled samples with low-confidence prediction, since we are unclear about their real labels, it is hard to generate mixed samples by them. In order to further improve the generalization ability of our method by fully leveraging these unlabeled samples, we employ entropy loss to encourage unlabeled samples to be classified into a specific category:
\begin{equation}
L_{ent} = \frac{1}{N_u}\sum_{i=1}^{N_u} H(F_c(F_g(x_i^u))),
\end{equation}
where $H(\cdot)$ is the entropy function. By introducing $L_{ent}$, the networks are forced to make more confident predictions on ambiguous unlabeled samples.

\subsection{Training Procedure}\label{training}

Formally, as aforementioned, we denote $\mathcal{S}_l$ as the set of all labeled samples, $\mathcal{S}_u$ as the set of all unlabeled samples. During training, once we select the reliable unlabeled samples with high confident pseudo-labels, forming $\mathcal{S}_{u'}$, we move them to the set of samples with pseudo-labels which is denoted as $\mathcal{S}_p$. At each training epoch, we assume, $|\mathcal{S}_l| = N_l$, $|\mathcal{S}_u| = N_u$ and $|\mathcal{S}_p| = N_p$. The total number of all training samples is denoted as $N= N_l + N_u + N_p$ (refer to Algorithm \ref{algo}).

For training our network, we apply 1) a classification loss $L_{cls}$ to $\mathcal{S}_l$ and $\mathcal{S}_p$ since they are with labels or predicted pseudo-labels, and 2) an adversarial loss $ L_{adv}$ to all sets of samples, {\em i.e.}, $\mathcal{S}_l$, $\mathcal{S}_p$, and $\mathcal{S}_u$.

\begin{algorithm}[t]
\KwIn{Labeled source $\mathcal{S}_l$, unlabeled source $\mathcal{S}_u$}
\KwOut{Generalizable model $F_g$ and $F_m$}
\textbf{Initialize} Networks, Pseudo-label Set $\mathcal{S}_p = \emptyset $, Class representation list $L = [\emph{none}]\times C$\ for each domain\;
\While{not end of epoch}{
$\mathcal{S}_m \leftarrow$ Perform domain mixup on $\mathcal{S}_l$ and $\mathcal{S}_p$\;
Training the model using Eqns. (\ref{twelve}) and (\ref{thirteen})\;

Inference the model on $\mathcal{S}_u$, and obtain $q(x^u)$\;

\If{no none in $L$}{
 $s(x^u)$ $\leftarrow$ Calculated by Eqn. (\ref{six})\;
}
Update class representation by $q(x^u)$\;
Assign pseudo-labels by $s(x^u)$\;
Recognize confident and ambiguous samples by Eqn. (\ref{pseu}), and the confident set is denoted as $\mathcal{S}_{u'}$\;
Update $\mathcal{S}_u \leftarrow \mathcal{S}_u-\mathcal{S}_{u'}$, $\mathcal{S}_p \leftarrow \mathcal{S}_p \cup \mathcal{S}_{u'}$;
}
\caption{Training Process}
\label{algo}
\end{algorithm}

\begin{equation}
 L_{cls} = \frac{1}{N_l}\sum_{i=1}^{N_l}  \ell(F_c(F_g(x_i^l)), {y_i^l}) + \frac{1}{N_p}\sum_{i=1}^{N_p}  \ell(F_c(F_g(x_i^p)), \widehat{y_i^p}),
\label{cls}
\end{equation}

$$
L_{adv}=\frac{1}{N}{(\sum_{i=1}^{N_l}\ell(F_d(F_g(x_i^l)),{z_i^l})}
\\+\sum_{i=1}^{N_u}\ell(F_d(F_g(x_i^u)),{z_i^u})$$
\begin{equation}
+\sum_{i=1}^{N_p}\ell(F_d(F_g(x_i^p)), {z_i^p})),
\end{equation}
where $\ell(\cdot,\cdot)$ is the cross-entropy loss. The loss on samples mixed up by $\mathcal{S}_p$ and $\mathcal{S}_l$ is defined as:

\begin{equation}
L_{cls\_ mix} = \frac{1}{N_l}\sum_{i=1}^{N_l} \sum_{j=1}^{N_p} \ell(F_m(F_g(\tilde{x})), \tilde{y}),
\end{equation}
\begin{equation}
L_{adv\_ mix} = \frac{1}{N_l}\sum_{i=1}^{N_l} \sum_{j=1}^{N_p} \ell(F_d(F_g(\tilde{x})), \tilde{z}).
\end{equation}
The training objective of our semi-supervised DG model can be described as follows:
\begin{equation}
\min _{F_g, F_c, F_m} L_{cls} + L_{cls\_mix} +w^t(-L_{adv} -L_{adv\_ mix} + L_{ent}),
\label{twelve}
\end{equation}
\begin{equation}
\min _{F_d}  L_{adv} + L_{adv\_ mix},
\label{thirteen}
\end{equation}
where the weight function $w^t$ ramps up from zero to one during the training procedure. The whole training procedure is described in Algorithm~\ref{algo}.

\section{Experiments}\label{s-experiment}

\subsection{Experimental Setting}\label{dataset}
\textbf{Datasets.}
To evaluate the effectiveness of our method for the semi-supervised domain generalization, we conduct several experiments on four benchmark datasets. \textbf{PACS} \cite{pacs} contains 7 categories of images from 4 domains (Photo, Art painting, Cartoon and Sketch). 
\textbf{OfficeHome} \cite{officehome} consists of images from 4 different domains (Artistic, Clip art, Product and Real-World). For each domain, this dataset involves 65 object categories found typically in office and home.
To verify that our method also has promotion when data is abundant, we validate our method on  $\textbf{miniDomainNet}$. It is a subset of DomainNet aggregated by \cite{minidomainnet}, which contains 140,006 images from 4 domains (Clipart, Painting, Real, and Sketch), covering 126 classes in the raw dataset. $\textbf{VLCS}$ \cite{vlcs} includes five categories of images from four domains (\ie, Caltech 101, PASCAL VOC, LabelMe and SUN09). For each dataset, we select two domains as the unlabeled source domains, one domain as the labeled source domain, and leave the remaining one for test. We test all 12 combinations of domains on each dataset and report the average accuracy.
To ensure a fair comparison with other methods, all experiments utilize the same scheme of data division.

\textbf{Implementation Details.}
In all experiments, we use ResNet \cite{resnet} as the backbone, and we start with a pre-trained model and fine-tune on source domains with the batch size of 128. Since the test domain is unavailable during training. For PACS and OfficeHome, we train 120 epochs and select the model obtained in the final epoch for test. For miniDomainNet, the network converges at an earlier epoch due to the huge amount of data. So we just train 60 epochs and save the final model.
We apply a SGD optimizer with momentum 0.9, and we set the initial learning rate to 1e-3 in the case of PACS, OfficeHome and miniDomainNet, and 1e-4 in the case of
VLCS. The learning rate is divided by 10 at the 30-th and the 50-th epochs. All experiments are implemented in Pytorch with $4\times 11$ GB RTX 2080Ti GPUs.

\subsection{Comparison with Other Methods}\label{cmp}
\renewcommand{\cmidrulesep}{0mm} 
\setlength{\aboverulesep}{0mm} 
\setlength{\belowrulesep}{0mm} 
\setlength{\abovetopsep}{0cm}  
\setlength{\belowbottomsep}{0cm}

On PACS, we compare our method with the following baselines and the state-of-the-art semi-supervised learning approaches (\ie, FixMatch \cite{fixmatch},  FeatMatch \cite{featmatch} and \rev{AdaMatch \cite{AdaMatch}}), domain adaptation (DA) approaches (\ie, SymNets \cite{symnets}, SRDC \cite{srdc}, CGDM \cite{cgdm}, \rev{ATDOC \cite{ATDOC} and FixBi \cite{FixBi}}), domain generalization (DG) methods (\ie, L2D \cite{singleDG}, JiGen \cite{JiGen} and RSC \cite{RSC}) and `DA+DG' methods on the classification accuracy of the target domain using ResNet-18 and ResNet-50. And we also 
compare ours with the baselines and the state-of-the-art methods on OfficeHome and miniDomainNet using ResNet-18. Finally, we report  our performance, the baseline results on VLCS using ResNet-18.

\begin{itemize}
\item \textbf{Baseline} 
    \begin{itemize}
        \item \textbf{SupOne:} Train a plain model on one labeled source domain in a supervised way and the other unlabeled source domains are not used.
        \item \textbf{DSDGN:} Implement the semi-supervised DG method proposed in \cite{semidgtwo} using the same network structure as our framework.
    \end{itemize}

\item \textbf{DA:} The labeled source domain is used as the source domain in the DA methods, and the mixture of the two unlabeled source domains is used as the target domain. Finally, the unseen target domain is used for testing.
\item \textbf{DG:} For single-DG methods, {\em i.e.} L2D, the unlabeled domains are not used. For DG methods, {\em i.e.} JiGen and RSC, unlabeled samples are used during training.

\item \textbf{DA+DG:} We consider the labeled domain as the source domain and other unlabeled domains as the unlabeled target domain. Then we use a UDA method to generate pseudo-labels for unlabeled domains. Finally, both the fully-labeled domain and pseudo-labeled unlabeled domains are fed into a supervised DG method to train the final model.
\end{itemize}

\begin{table*}[t]
  \centering
  \caption{
Experimental results of accuracy (\%) on PACS based on ResNet-18 and ResNet-50. The title in the first row indicates the name of the target domain and the title in the second row is the name of the labeled domain among training source domains. \textbf{P}, \textbf{A}, \textbf{C}, \textbf{S} are Photo, Art painting, Cartoon, Sketch, respectively. Note that the best performance is in \textbf{bold}.}

  \resizebox{\textwidth}{50mm}{
    \begin{tabular}{|cc|ccc|ccc|ccc|ccc|c|}
    \hline
    \multicolumn{2}{|c|}{\multirow{2}[1]{*}{Method}} & \multicolumn{3}{c|}{Photo} & \multicolumn{3}{c|}{Art Painting} & \multicolumn{3}{c|}{Cartoon} &
    \multicolumn{3}{c|}{Sketch} &
    \multirow{2}[1]{*}{Avg.} \\
&& A & C & S & P & C & S & P & A & S & P & A & C &  \\
\hline
\multicolumn{15}{|c|}{ResNet-18} \\
    \hline
    \multirow{2}[1]{*}{Baseline} & \multicolumn{1}{|c|}{SupOne} & 95.69 & 86.17 & 40.66 & 64.60 & 68.75 & 25.93 & 25.90 & 58.53 & 38.10 & 32.55 & 49.05 & 61.29 & 53.94 \\
  & \multicolumn{1}{|c|}{DSDGN \cite{dsdgn}} & \textbf{96.29} & 88.38 & 36.94 & 66.46 & 70.36 & 26.12 & 47.27 & 67.96 & 50.64 & 46.35 & 59.35 & 61.77 & 59.82 \\
  \hline
  \multirow{3}[1]{*}{SSL} & \multicolumn{1}{|c|}{FixMatch \cite{fixmatch}} & 95.39 & 85.63 & 59.40 & 65.38 & 68.41 & 51.37 & 45.22 & 62.20 & 55.55 & 53.32 & 61.52 & \textbf{76.94} & 65.03 \\
  & \multicolumn{1}{|c|}{FeatMatch \cite{featmatch}}& 95.33 & 82.57 & 56.89 & 66.06 & 72.02 & 58.69 & 47.10 & 65.57 & 57.30 & 64.39 & 72.59 & 74.40 & 67.74 \\
  & \multicolumn{1}{|c|}{AdaMatch \cite{AdaMatch}}& 94.61 &	49.64 &	41.20 &	69.43 &	\textbf{81.01} 	&42.48 	& 64.12 &65.53 	&60.28& 	54.38 &	68.30 &	76.27 	&63.94  \\
  \hline
  
  \multirow{3}[1]{*}{DG} &  \multicolumn{1}{|c|}{L2D \cite{singleDG}} & 95.51	& 86.65	& 47.25 &	64.75	& 73.78	& 49.95	& 38.35	& 68.77	& 63.31	& 41.23	& 67.24	& 70.20	& 63.92 \\
  & \multicolumn{1}{|c|}{JiGen \cite{JiGen}}& 95.39	& 83.53	& 47.43	& 66.26	& 69.53	& 33.54	& 34.68	& 66.98	& 54.18	& 40.83	& 57.78	& 62.61	& 59.40 \\
  & \multicolumn{1}{|c|}{RSC \cite{RSC}} & 86.23	& 84.79	& 46.05	& 63.38	& 68.75	& 33.15
	& \textbf{66.42}	& 47.06	& 57.34	& \textbf{66.15} & 70.83	& 60.09	& 62.52 \\
	\hline
  \multirow{5}[1]{*}{DA} &  \multicolumn{1}{|c|}{SymNets \cite{symnets}} & 88.42	& 76.44 & 28.36 &	61.80	& 48.27 &	31.86 &	30.21 &	\textbf{74.24} &	45.69 &	16.36 &	53.41 &	58.70 &	51.15 \\
  & \multicolumn{1}{|c|}{SRDC \cite{srdc}}& 91.50 &	80.36	& 38.32	& 55.37	& 71.68 &	31.40 &	50.43 &	69.28 &	54.73 &	21.41 &	54.75 &	60.15 &	56.62 \\
  & \multicolumn{1}{|c|}{CGDM \cite{cgdm}} & 95.15 &	75.69 &	47.01 &	62.26 &	64.86 &	31.69 &	39.38 &	59.56 &	50.73 &	12.57 &	41.84 &	59.86 &	53.38 \\
  & \multicolumn{1}{|c|}{ATDOC \cite{ATDOC}} & 90.06 & 78.86 &	\textbf{79.52} &	56.20 &	53.52 &	46.78 &	57.68 	&60.54 &	55.16 &	46.63 &	51.01 &	42.30 &	59.86  \\
  & \multicolumn{1}{|c|}{FixBi \cite{FixBi}} & 87.90 & 	85.75 	&48.02 &	45.41 	&68.51 &	47.95 &	39.38 &	62.07 &	50.51 &	27.46 &	40.85 &	52.97 &	54.73  \\
  
  \hline
  \multirow{5}[1]{*}{DA+DG} &  \multicolumn{1}{|c|}{SymNets+RSC} & 92.28 &	\textbf{92.46} &	40.90 &	69.46 &	72.87 &	46.70 &	60.96 &	30.59 &	51.28 &	63.78 &	\textbf{73.05} &	18.91 &	59.44 \\
  & \multicolumn{1}{|c|}{SRDC+RSC} & 92.46 & 91.74 &	50.36 &	66.36 &	70.61 &	\textbf{65.87} &	58.23 &	57.21 &	\textbf{65.49} &	64.98 &	66.38 &	62.26 &	67.66  \\
  & \multicolumn{1}{|c|}{CGDM+JiGen} & 93.87 & 91.66	& 58.54 &	65.75 &	75.17 &	58.86 &	46.01 &	72.54 &	54.54 &	61.40 &	68.57 &	61.98 &	67.41 \\
  & \multicolumn{1}{|c|}{ATDOC+RSC} & 92.81 &	73.41 &	77.01 &	\textbf{70.46} &	61.28 &	59.81 &	60.20 &	65.61 &	64.29 &	49.55 &	65.64 &	72.18 &	67.69   \\
  
  & \multicolumn{1}{|c|}{FixBi+RSC} & 92.99 &	87.31 &	50.48 &	66.46 &	73.24 &	57.67 &	40.19 &	62.63 &	53.67 &	26.22 &	60.96 &	70.25 &	61.84  \\
  \hline
  & \multicolumn{1}{|c|}{Ours} & 94.37 & 91.02 & 66.53 & 69.92 & 75.68 & 55.37 & 54.22 & 71.46 & 57.94 & 65.69 & 71.27 & 71.06 & \textbf{70.38} \\
 \hline
 \multicolumn{15}{|c|}{ResNet-50} \\
 \hline
   \multirow{2}[1]{*}{Baseline} & \multicolumn{1}{|c|}{SupOne} & \textbf{98.14} & 86.71 & 36.89 & 73.39 & 72.46 & 30.37 & 34.47 & 65.87 & 45.86 & 34.03 & 56.76 & 68.39 & 58.61\\
   & \multicolumn{1}{|c|}{DSDGN \cite{dsdgn}} & 97.90 & 93.65 & 38.62 & 71.88 & 79.15 & 39.60 & 45.95 & 70.78 & 53.11 & 41.46 & 68.36 & 70.12 & 64.22\\
   \hline
  \multirow{3}[1]{*}{SSL} & \multicolumn{1}{|c|}{FixMatch \cite{fixmatch}} & 95.59 & 94.07 & 58.04 & 68.16 & 85.28 & 55.90 & 54.47 & 76.58 & 66.55 & 46.09 & 73.40 & 77.70 & 70.99\\
  & \multicolumn{1}{|c|}{FeatMatch \cite{featmatch}} & 97.84 & 93.71 & 52.22 & 74.37 & 86.87 & 58.74 & 66.13 & 71.08 & 65.78 & \textbf{71.54} & 73.84 & \textbf{81.04} & 74.43 \\
   & \multicolumn{1}{|c|}{AdaMatch\cite{AdaMatch}} & 96.95 &	93.29	&50.84	&71.29&	\textbf{88.13} &	48.39&	67.02	&66.17	&59.98	&62.65	&67.54&	77.32	&70.80 \\
  \hline
  \multirow{3}[1]{*}{DG} & \multicolumn{1}{|c|}{L2D \cite{singleDG}} & 97.13	& 92.28	& 52.04	& 73.63	& 78.08	& 52.10	& 43.30	& 74.40	& 69.75	& 44.08	& 66.48	& 76.18	& 68.29 \\
  & \multicolumn{1}{|c|}{JiGen \cite{JiGen}} & 97.31	& 89.76	& 53.23	& 72.27	& 76.27	& 42.04	& 42.28	& 70.01	& 64.59	& 45.71	& 59.05	& 69.00	& 65.13 \\
  & \multicolumn{1}{|c|}{RSC \cite{RSC}}  &	
  91.56	& 88.80 &	59.10 &	64.31 &	68.70 &	34.42 &	\textbf{68.56}	& 47.78 &	65.49 &	65.46 &	 74.12 &	62.13 &	65.87 \\
  \hline
   \multirow{5}[1]{*}{DA} &  \multicolumn{1}{|c|}{SymNets \cite{symnets}} &  91.52	& 75.13 &	52.99 &	64.77 &	60.84 &	38.97 &	38.67 &	67.41	& 51.49 &	29.55 &	56.71 &	62.41 &	57.54\\
  & \multicolumn{1}{|c|}{SRDC \cite{srdc}}&  92.52 &	93.89 &	31.74 &	65.14 &	77.98 &	42.63 &	44.45 &	72.40 &	57.85 &	24.33 &	53.09 &	60.22 &	59.69\\
  & \multicolumn{1}{|c|}{CGDM \cite{cgdm}} & 97.37	& 76.35	& 69.46	& 72.27	& 64.01	& 54.10	& 51.02	& 62.03	& 57.51	& 28.94	& 43.15	& 63.60	& 61.65 \\
   &  \multicolumn{1}{|c|}{ATDOC\cite{ATDOC}} & 92.87 	&82.81 &	88.38 &	39.70 &	76.07 & 	55.86 &	55.50 &	68.00 &	57.68 	&38.69 &	52.00 &	53.53 	&63.42  \\
  & \multicolumn{1}{|c|}{FixBi \cite{FixBi}}&  90.24 &	92.10 &	55.27 &	64.21 &	74.51 &	52.05 &	37.59 &	70.99 &	65.10 &	45.48 &	54.57 &	60.88 &	63.58 \\
   \hline
   \multirow{5}[1]{*}{DA+DG} & 
  \multicolumn{1}{|c|}{SymNets+RSC}&  94.79 &	94.56 &	40.27 &	77.87 &	74.52 &	50.18 &	61.12 &	44.06 &	54.93 &	62.24 &	\textbf{76.99} &	30.25 &	 63.48 \\
      & \multicolumn{1}{|c|}{SRDC+RSC}&  95.93 &	\textbf{95.03} &	64.25 &	75.39 &	78.93 &	60.40 &	59.60 &	74.27 &	67.41 &	58.18 &	72.10 &	74.40 &	72.99 \\
  & \multicolumn{1}{|c|}{CGDM+JiGen} & 97.60 &	94.91 &	73.03 &	78.02 &	83.89 &	73.44 &	49.87 &	64.97 &	63.35 &	57.04 &	64.72 & 73.63 &	72.87 \\
  & \multicolumn{1}{|c|}{ATDOC+RSC} &95.57 &	80.30 &	\textbf{90.06} &	62.30 &	81.15 	&\textbf{76.81}& 	58.32& 	72.99& 	\textbf{74.87} &	51.06 &	71.06 &	78.37 &	74.41  \\
  
  & \multicolumn{1}{|c|}{FixBi+RSC} & 95.99 &	93.35 &	54.79 &	69.97 &	77.88 &	55.57 &	35.07 &	70.99 &	67.41 &	67.29 &	71.37 &	75.06 &	69.56    \\
   \hline
   & \multicolumn{1}{|c|}{Ours} & 97.61 & 93.53 & 66.35 & \textbf{78.03} & 86.98 & 62.45 & 59.17 & \textbf{76.88} & 69.37 & 65.61 & 74.09 & 78.52 & \textbf{75.72} \\
    \hline
    \end{tabular}
    }
  \label{pacsresults}
\end{table*}

Table~\ref{pacsresults} displays the results on PACS. Compared with the baselines, our method achieves outstanding performance by significant margins with both smaller and larger network architectures. Specifically, our method improves the performance of SupOne by +16.44\% and +17.11\% with ResNet-18 and ResNet-50, respectively. This shows that ours significantly improves the performance by leveraging unlabeled data. Compared with DSDGN, ours gains +10.56\% with ResNet-18 and gains +11.50\% with ResNet-50. The results indicate that our method makes more efficient use of unlabeled samples, thus our proposed domain-aware pseudo-labeling and other modules outperform the naive pseudo-labeling method in DSDGN.

Compared with the recent SOTA semi-supervised methods, ours also shows priority in the SSDG task. Specifically, ours outperforms the best SSL methods by +2.64\% using ResNet-18 and +1.29\% using ResNet-50. This shows that conventional semi-supervised methods are not superior in solving SSDG problems. By considering the domain shift between the labeled source domain and unlabeled source domains, our proposed method can achieve better performance in a clear margin compared with semi-supervised approaches.

From Table \ref{pacsresults} we also observe that using ResNet-18 as a backbone architecture, \rev{ours outperforms all DA methods ({\em i.e.}, SymNets, SRDC, CGDM, ATDOC and FixBi) by large margins: +19.23\%, +13.76\%, +17.00\%, +10.52\% and +15.65\%.} Ours is also clearly better than DA methods when ResNet-50 is employed as a backbone architecture. This main reason is that these DA methods do not address the domain gap between source domains and the unseen target domain because the unseen target domain cannot be employed during training in the SSDG task.

Additionally, as seen in Table \ref{pacsresults}, our method achieves better performance than both single-DG and DG methods. For example, our method improves L2D by +6.46\% using ResNet-18 and +7.43\% using ResNet-50. These results show that our method improves the performance of single-DG with free unlabeled data. Ours also outperforms the best DG methods by +7.86\% using ResNet-18 and +9.85\% using ResNet-50. These results imply that our method makes more efficient use of unlabeled samples, \ie, the proposed domain-aware pseudo-labeling module and the proposed dual-classifier surpass ``self-challenging" mechanism (RSC) and ``solving a jigsaw puzzle" task (JiGen) in utilizing unlabeled samples.

Moreover, we can observe in Table \ref{pacsresults} that compared with the SOTA `DA+DG' methods using ResNet-18 as backbone, ours shows advantages. Specifically, compared with `SymNets+RSC', our method improves the average accuracy by +10.94\%. Compared with `SRDC+RSC', ours gains +2.72\%, and compared with `CGDM+JiGen', there is an improvement of +2.97\%. \rev{Compared with `ATDOC+RSC', ours improves +2.69\%, and compared with `FixBi+RSC', ours gains +8.54\%.}
Moreover, using ResNet-50 as backbone,  \rev{compared with `SymNets+RSC', `SRDC+RSC', `CGDM+JiGen', `ATDOC+RSC' and `FixBi+RSC', ours improves the accuracy by +12.24\%, +2.73\%, +2.85\%, +1.31\% and +6.16\%, respectively.} The reason why `DA+DG' methods are not superior to ours is that `DA+DG' methods are not an end-to-end deep framework, thus the labeling process of UDA methods cannot be improved by the process of DG, which means that the performance of DG is largely influenced by that of UDA. Particularly, when the UDA model is unreliable, the performance of the whole model would become inferior.

\begin{table*}[t]
  \centering
  \caption{
Experimental results of accuracy (\%) on OfficeHome. The title in the first row indicates the name of the target domain and the title in the second row is the name of the labeled domain in training source. \textbf{A}, \textbf{C}, \textbf{P}, \textbf{R} stand for Art, Clipart, Product, Real, respectively. The best performance is \textbf{bold}.}

  \resizebox{\textwidth}{20mm}{
    \begin{tabular}{|cc|ccc|ccc|ccc|ccc|c|}
    \hline
    
    \multicolumn{2}{|c|}{\multirow{2}[1]{*}{Method}} & \multicolumn{3}{c|}{Art} & \multicolumn{3}{c|}{Clipart} & \multicolumn{3}{c|}{Product} &
    \multicolumn{3}{c|}{Real} &
    \multirow{2}[1]{*}{Avg.} \\

&& C & P & R & A & P & R & A & C & R & A & C & P &  \\
\hline

    \multirow{2}[1]{*}{Baseline} & \multicolumn{1}{|c|}{SupOne} & 43.63 & 38.20 & 54.84 & 38.95 & 37.34 & 44.10 & 54.88 & 54.02 & 70.94 & 64.08 & 57.54 & 62.98 & 51.79 \\
  & \multicolumn{1}{|c|}{DSDGN \cite{dsdgn}} & 45.49 & 41.37 & 56.49 & 38.40 & 37.46 & 43.98 &  55.37 & 54.99 & 72.11 & 64.56 & 55.93 & 63.44 & 52.47\\
  \hline
  \multirow{2}[1]{*}{SSL} 

    & \multicolumn{1}{|c|}{FixMatch \cite{fixmatch}}& 44.53 & 42.14 & 58.11 & 43.42 & 42.09 & 45.67 & 56.24 & 56.39 & 70.80 & 65.38 & 56.43 & \textbf{65.71} & 53.91 \\
  
  & \multicolumn{1}{|c|}{FeatMatch \cite{featmatch}}& 49.98 & 47.34  & 55.67 & 41.95  & 42.96 & 47.33 & 54.11 & 53.71 & 68.19 & 61.12 &  54.99 & 64.33 & 53.47 \\

  \hline
  
  \multirow{3}[1]{*}{DG} &  \multicolumn{1}{|c|}{L2D \cite{singleDG}} & 42.14 & 38.25 & 56.72  & 40.75 & 41.42 & \textbf{51.33} & 47.04  & 51.78 & 69.90 & 56.90 &  54.57 & 63.58 & 51.20\\

  &  \multicolumn{1}{|c|}{JiGen \cite{JiGen}} & 32.06 & 30.49 & 42.85 & 35.51 & 34.71 & 41.70 & 42.60 & 47.38 & 62.78 & 52.51 & 49.28 & 54.26 & 43.84 \\

  & \multicolumn{1}{|c|}{RSC \cite{RSC}} & 40.08 & 37.19 & 53.99 & 39.56 & 38.37 & 45.43 & 48.13 & 57.12 & 69.45 & 63.46 & 52.05 & 60.80 & 50.47 \\

	\hline
  \multirow{3}[1]{*}{DA} 

  & \multicolumn{1}{|c|}{SymNets \cite{symnets}}& 39.64 & 45.20 & 53.65& 33.29 & 32.30 & 37.96 & 50.96 & 48.52 & 68.17& 58.73 & 51.11& 60.12 & 48.30 \\
  
  & \multicolumn{1}{|c|}{SRDC \cite{srdc}}& \textbf{50.31} & \textbf{52.37} & \textbf{58.96} 	& 37.18	& 37.27 & 38.24	 & 58.68	 &	57.87 &	68.35 & 61.26	 & 59.08	& 64.61	& 53.68	 \\

  & \multicolumn{1}{|c|}{CGDM \cite{cgdm}}& 46.14 &44.91&51.92&38.01& 37.69 &42.97&57.22&58.43&69.43&64.86 & 59.63 & 63.69 & 52.91	 \\
  \hline
  \multirow{3}[1]{*}{DA+DG} 

  & \multicolumn{1}{|c|}{SymNets+RSC} &  30.74 &37.54&51.76&33.24&39.25&43.76&42.69&46.18&65.71&49.02 & 41.27 & 60.16 & 45.11 \\
  & \multicolumn{1}{|c|}{SRDC+RSC} & 43.51 & 45.74 & 49.40	 &	39.43 & \textbf{43.30} & 39.40	 & 54.04	 & 	55.64 & 69.09	 &	56.92 &	55.73 &	 63.35&	51.30  \\
  & \multicolumn{1}{|c|}{CGDM+JiGen} & 49.36 & 48.66 & 53.59 &41.28 & 39.52&43.43& \textbf{60.35} & \textbf{61.63} &71.09 & 63.21 & \textbf{61.35} & 64.03& 54.79  \\

  \hline
  & \multicolumn{1}{|c|}{Ours} &  47.55 & 46.07 & 58.01 & \textbf{44.33} & 42.34 & 47.90 & 57.56 & 57.83 & \textbf{72.43} & \textbf{65.48} & 59.74 & 65.09 & \textbf{55.36} \\
    \hline
    \end{tabular}
    }
  \label{officehomeresults}
\end{table*}

\begin{table*}[t]
  \centering
  \caption{Experimental results of accuracy (\%) on miniDomainNet. The title in the first row indicates the name of the target domain, and the title in the second row is the name of the labeled domain in source domains. \textbf{C}, \textbf{P}, \textbf{R} and \textbf{S} stand for Clipart, Painting, Real and Sketch, respectively. The best performance is \textbf{bold}.}

  \resizebox{\textwidth}{20mm}{
    \begin{tabular}{|cc|ccc|ccc|ccc|ccc|c|}
    \hline
    
    \multicolumn{2}{|c|}{\multirow{2}[1]{*}{Method}} & \multicolumn{3}{c|}{Clipart} & \multicolumn{3}{c|}{Painting} & \multicolumn{3}{c|}{Real} &
    \multicolumn{3}{c|}{Sketch} &
    \multirow{2}[1]{*}{Avg.} \\

&& P & R & S & C & R & S & C & P & S & C & P & R &  \\
\hline

    \multirow{2}[1]{*}{Baseline} & \multicolumn{1}{|c|}{SupOne} & 47.45 & 47.83 & 53.54 & 37.91 & 48.43 & 42.69 & 48.53 & 64.90 & 47.35 & 43.21 & 42.12 & 42.33 & 47.19 \\
  & \multicolumn{1}{|c|}{DSDGN \cite{dsdgn}} & 47.29 & 48.16 & 53.25 & 39.27& 50.88 & 46.34 & 44.78 & 64.46 & 50.21 & 42.18 & 44.50 & 42.61 & 47.83\\

  \hline
  \multirow{2}[1]{*}{SSL} 
  & \multicolumn{1}{|c|}{FixMatch \cite{fixmatch}}& 50.79 & 56.89 &57.46 & 48.75 & 60.38 & 57.18 &52.47 & 63.08&56.54&38.09&42.22& 40.20 &52.00\\
  
  & \multicolumn{1}{|c|}{FeatMatch \cite{featmatch}}& 52.70 &  52.17 & 52.06 &  \textbf{48.91} & 53.38 & 37.05 & 37.40 & \textbf{66.06} & 41.14 & 45.57 & \textbf{49.43}  & 47.17 & 48.59 \\
  \hline
  
  \multirow{3}[1]{*}{DG} &  \multicolumn{1}{|c|}{L2D \cite{singleDG}} & 50.23 & 52.75 & 53.90  & 37.47 & 
  \textbf{56.49} & 42.40& 45.69  & 59.16 & 45.81 & 46.31 &  45.59 & 45.37 & 48.43\\

  & \multicolumn{1}{|c|}{JiGen \cite{JiGen}} & 42.38 & 46.87 & 44.27& 27.81 & 43.22 & 26.23 & 37.39 & 54.01 & 37.44 & 33.45 & 30.18 & 33.06 & 38.03 \\
  
  & \multicolumn{1}{|c|}{RSC \cite{RSC}} & 42.41 & 48.20 & 47.84 & 28.00 & 52.65 & 33.07 & 37.63 & 56.48 & 35.49  &  41.92 & 42.68 & 40.59 & 42.25 \\

	\hline
  \multirow{3}[1]{*}{DA} 

& \multicolumn{1}{|c|}{SymNets \cite{symnets}}& 39.51&50.76&47.73&35.11&52.35&46.10&43.42&58.95& 48.48 & 37.14 & 33.44	&32.21 & 43.77\\
  & \multicolumn{1}{|c|}{SRDC \cite{srdc}}& 46.40 & 46.64 & 52.00 &42.18&52.11&50.53&55.71 & 65.49 &56.06&41.75&39.91 & 38.30 & 48.92	 \\
  & \multicolumn{1}{|c|}{CGDM \cite{cgdm}}& 50.31 & 49.71 &  56.03	& 42.64	& 53.41 & 48.57	 & 52.27	 & 64.52	 & 54.76 & 43.39	 & 40.70	& 42.20	& 49.88	 \\
  \hline
  \multirow{3}[1]{*}{DA+DG} 
  & \multicolumn{1}{|c|}{SymNets+RSC} & 47.66&48.09&47.24& 39.21 &51.01& 52.28 & 40.14 & 52.57 & 49.68 & 38.80 & 40.17 & 39.04 &  45.49 \\
  & \multicolumn{1}{|c|}{SRDC+RSC} & 48.74&45.86&55.25& 41.12&55.37& 52.23 & 48.23 &59.33&52.94& 39.73 & 43.34 & 43.04 & 48.77\\
  & \multicolumn{1}{|c|}{CGDM+JiGen} & 49.79 & \textbf{58.34} & 56.75	 &	45.20 & 54.73 & 51.91	 & 	\textbf{55.76} & 62.19	 & \textbf{58.77}	 &	46.96 & 45.83 &	46.27& 52.71  \\
  \hline
  & \multicolumn{1}{|c|}{Ours} & \textbf{54.29} & 55.98 & \textbf{58.16} & 45.36 & 56.22 & \textbf{53.08} & 51.51 & 64.55 & 55.05 & \textbf{48.70} & 49.38 & \textbf{49.91} & \textbf{53.52} \\
    \hline
    \end{tabular}
    }
  \label{minidomainnetresults}
\end{table*}

\begin{table*}[t]
  \centering
  \caption{Experimental results of accuracy (\%) on VLCS. The title in the first row indicates the name of the target domain and the title in the second row is the name of the labeled domain among training source domains. C, L, V, S stand for CALTECH, LABELME, VOC, SUN, respectively. The best performance is in \textbf{bold}.}
  \resizebox{\textwidth}{11mm}{
    \begin{tabular}{|c|ccc|ccc|ccc|ccc|c|}
    \hline
    \multirow{2}[1]{*}{Method} & \multicolumn{3}{c|}{CALTECH} & \multicolumn{3}{c|}{LABELME} & \multicolumn{3}{c|}{VOC} &
    \multicolumn{3}{c|}{SUN} &
    \multirow{2}[1]{*}{Avg.} \\
& L & V & S & C & V & S & C & L & S & C & L & V &  \\
\hline
    SupOne &  68.83	& 98.02 &	72.23	& 53.09 &	57.27 &	59.64 &	44.52 &	59.87 &	61.37	& 35.44 &	\textbf{50.24} &	73.43 &	61.16 \\
  DSDGN\cite{dsdgn} & 70.18 &	\textbf{98.30} &	\textbf{74.35} &	48.01 &	65.85 &	59.22 &	49.01 &	56.40 &	61.08 &	38.73 &	\textbf{50.24} &	70.35 &	61.81\\
 Ours & \textbf{79.01} &	97.39 &	72.93 &	\textbf{65.85} &	\textbf{66.08} &	\textbf{61.11} &	\textbf{51.08} &	\textbf{60.38} &	\textbf{63.12} &	\textbf{39.64} &	49.30	& \textbf{74.40} &	\textbf{65.02} \\
   
    \hline
    \end{tabular}
    }
  \label{vlcs_results}
\end{table*}

Experimental results on OfficeHome are shown in Table~\ref{officehomeresults}. It is worth noting that OfficeHome has a relatively smaller domain shift than PACS. As seen in this table, our method gains +3.57\% compared with SupOne and improves the average accuracy by +2.89\% compared with DSDGN, which thanks to the efficacy of the proposed domain-aware pseudo-labeling scheme and dual-classifier. Furthermore, compared with the SOTA SSL, DG, DA and `DA+DG' methods, our method increases \rev{+1.45\%}, +4.16\%, +1.68\% and +\rev{0.57\%}, respectively.

We compare our method with baselines and the SOTA methods on miniDomainNet in Table~\ref{minidomainnetresults}. As seen, our method achieves +6.33\% and +5.69\% gain on miniDomainNet compared with SupOne and DSDGN, respectively. 
Besides, our method outperforms all the SOTA methods (\ie, SSL, DG, DA and `DA+DG') by large margins: \rev{+1.52\%}, +5.09\%, +3.64\% and +0.81\%. 
The results indicate that our method is also effective on the large-scale dataset.

In Table \ref{vlcs_results} we report the performance of ours and baselines on VLCS. We notice that SupOne achieves an average accuracy of 61.16\%, which is only trained on the labeled source domain in a supervised way. DSDGN achieves slight improvement compared with SupOne. Our method increases the average accuracy by +3.86\% compared with SupOne on VLCS, which could be attributed to that we assign accurate pseudo-labels to unlabeled source domains and the inter-domain mixup improves the generalization ability of the model.

\subsection{Ablation Study} \label{s-ablation}

\begin{figure}[t]
\centering
\includegraphics[width=0.7\columnwidth]{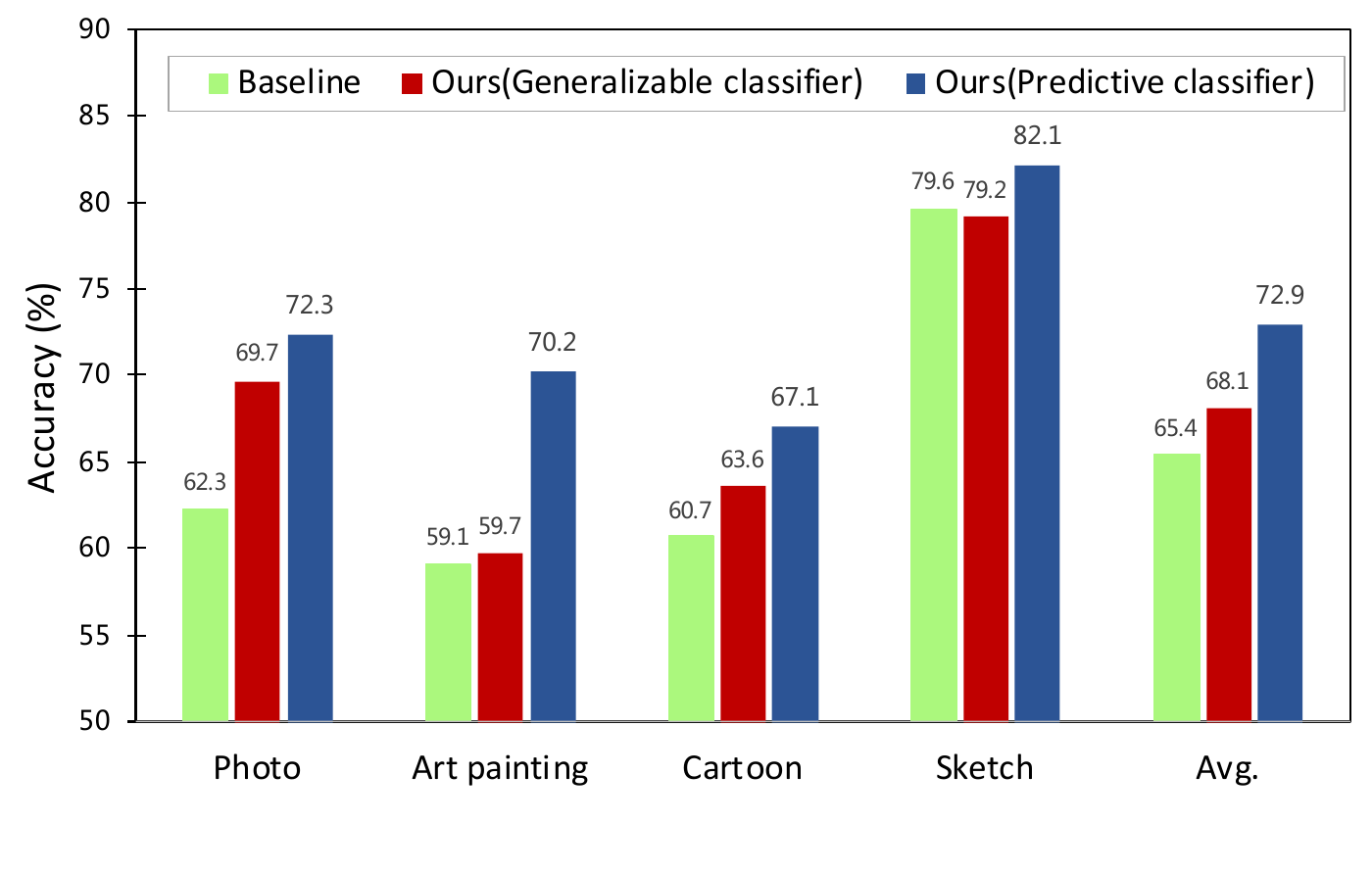} 
\caption{Pseudo-label accuracy (\%) on PACS.}
\label{pesudoacc}
\end{figure}

\begin{table*}[htbp]
  \centering
  \caption{Accuracy (\%) of ablation study on PACS. \textbf{P}, \textbf{A}, \textbf{C}, \textbf{S} stand for Photo, Art painting, Cartoon, Sketch, respectively. Baseline represents our method without both DAPL and DC. The best performance is \textbf{bold}.}
  \resizebox{\textwidth}{15mm}{
    \begin{tabular}{|c|ccc|ccc|ccc|ccc|c|}
    \hline
    \multirow{2}[1]{*}{Method} & \multicolumn{3}{c|}{Photo} & \multicolumn{3}{c|}{Art Painting} & \multicolumn{3}{c|}{Cartoon} &
    \multicolumn{3}{c|}{Sketch} &
    \multirow{2}[1]{*}{Avg.} \\
    
& A & C & S & P & C & S & P & A & S & P & A & C &  \\
\cmidrule{1-14}
        
    Baseline & 92.52 & 87.49 & 48.80 & 66.80 & 72.51 & 45.51 & 51.07 & 67.53 & 51.83 & 52.07 & 53.47 & 59.36 & 62.41 \\
  
  Ours w/o DAPL & 92.64 & 88.38 & 49.40 & 69.24 & 69.14 & 45.02 & 54.10 & 68.94 & 53.41 & 58.77 & 64.01 & 65.20 & 64.85 \\
  
  Ours w/o DC & 93.29 & 85.69 & 60.96 & 66.31 & 73.68 & 46.83 & \textbf{56.83} & 68.13 & 52.30 & 54.26 & 56.05 & 63.32 & 64.80 \\

  DBSCAN\cite{dbscan} & 93.47 & 85.57 & 57.78 & 66.21
& 70.07 & 43.07 & 49.62 & 67.19 &   48.81 & 44.31 & 54.11 & 54.42 & 61.22 \\
  
  Agglomerative\cite{agglo} & 85.81 & 89.28 & 62.28
& 65.82 & 69.63 & 52.59 & 51.07 & 67.66
& 50.77 & 46.02 & 56.15 & 56.07 & 62.76 \\
  
  Ours & \textbf{94.37} & \textbf{91.02} & \textbf{66.53} & \textbf{69.92} & \textbf{75.68} & \textbf{55.37} & 54.22 & \textbf{71.46} & \textbf{57.94} & \textbf{65.69} & \textbf{71.27} & \textbf{71.06} & \textbf{70.38} \\
    
    \hline
    \end{tabular}%
    }
  \label{ablation}%
\end{table*}%
In order to verify the contributions of domain-aware pseudo-labeling (DAPL) and dual-classifier (DC), we conduct an ablation study on PACS, as reported in Table~\ref{ablation}.
``Ours w/o DAPL'' replaces DAPL by naive pseudo-labeling, and ``Ours w/o DC'' conducts prediction and generalization by the same classifier in the training stage.
Without DAPL or DC, the generalization ability of our method dramatically drops. 
The results show that our modules are crucial to improving the accuracy of the classification on unseen target domains. 
Besides, we show the pseudo-label accuracy in Figure~\ref{pesudoacc}. As seen, our method can obtain more accurate pseudo-labels for unlabeled data compared with the baseline (\ie, our method removes both DAPL and DC). 
In order to further evaluate the effectiveness of DAPL, we conduct an ablation study to replace DAPL with other state-of-the-art cluster methods (  \ie, DBSCAN~\cite{dbscan} and Agglomerative Clustering~\cite{agglo}) for comparison. As seen, ours outperforms the best clustering-based methods by +7.62\%. The results indicate that domain-aware pseudo-labeling is more effective compared with clustering methods.

In addition, as mentioned above, we propose to utilize a dual-classifier (a predictive classifier for producing pseudo-labels and a generalizable classifier) to avoid the possible accuracy degradation of pseudo-labels, which leverages the independent classifier for joint pseudo-label assignment and domain generalization. The predictive classifier is trained by original samples and the generalizable classifier is trained by mixed samples.
We also compare the pseudo-label accuracy of the two classifiers on PACS in Figure~\ref{pesudoacc}. As seen, the predictive classifier gains 4.8\% in average accuracy compared with the generalizable classifier. This shows that the generalizable model causes a drop in the accuracy of pseudo-labels. And the dual-classifier module is effective for mitigating this problem. It is worth mentioning that our method picks up pseudo labels with an accuracy of over 70\% in a 7-class classification task using ResNet-18.

\begin{table}[tbp]
  \centering
  \caption{Experimental results of accuracy(\%) with different designs of Domain Mixup and different loss functions on PACS. The titles indicate the target domains, and each column is the average accuracy of three combinations of source domains (\eg, A, C, S under Photo).}
    \begin{tabular}{|c|c|c|c|c|c|}
    \hline
Target & Photo & Art & Cartoon & Sketch & Avg. \\
    \midrule
 w/o Mixup & 80.48 & 63.48 & 62.88 & 60.03& 66.72 \\
 MixupAll & 79.50 & 63.51 & 59.24 & 63.16 & 66.35 \\
 
 w/o $L_{cls\_mix}$ & 77.61 & 61.63 & 57.95 & 55.38 & 63.14\\ 
 w/o $L_{adv}$ & 79.44 & 65.90 & 59.88 & 55.15 & 65.09\\ 
w/o $L_{adv\_mix}$ & 81.66 & 67.58 & 61.96 & 64.87 & 69.02\\ 
 w/o $L_{ent}$ & 82.69 & 62.44 & 59.43 & 67.15 & 67.93 \\ 
 Ours & 83.97 & 66.99 & 61.21 & 69.34 & 70.38 \\
 
    \hline
    \end{tabular}
  \label{ablationmixup}
\end{table}

As described in Sec. \ref{mining}, domain mixup is applied to promote the generalization ability of our model on the unseen target domain. And we only choose confident unlabeled samples to mix up with the labeled ones. We study the influence of different designs and show the results in Table \ref{ablationmixup}. As seen, our method shows better performance compared with ``w/o mixup'' and the model that mixes up all unlabeled samples. We also perform ablation study on the losses in the training objective, \ie, $L_{cls\_mix}$, $L_{adv}$, $L_{adv\_mix}$ and $L_{ent}$. The results show that ours outperforms the scheme without $L_{cls\_mix}$, $L_{adv}$, $L_{adv\_mix}$ and $L_{ent}$ by 7.24\%, 5.29\% , 1.36\% and 2.45\%, respectively, demonstrating the effectiveness of each loss function. In summary, in these experiments, we confirm that the main modules in our framework are useful.

\subsection{Further Analysis}\label{analysis}

\begin{figure}[tbp]
\centering
\subfigure{
\begin{minipage}[t]{0.48\linewidth}
\centering
\includegraphics[width=2.3in]{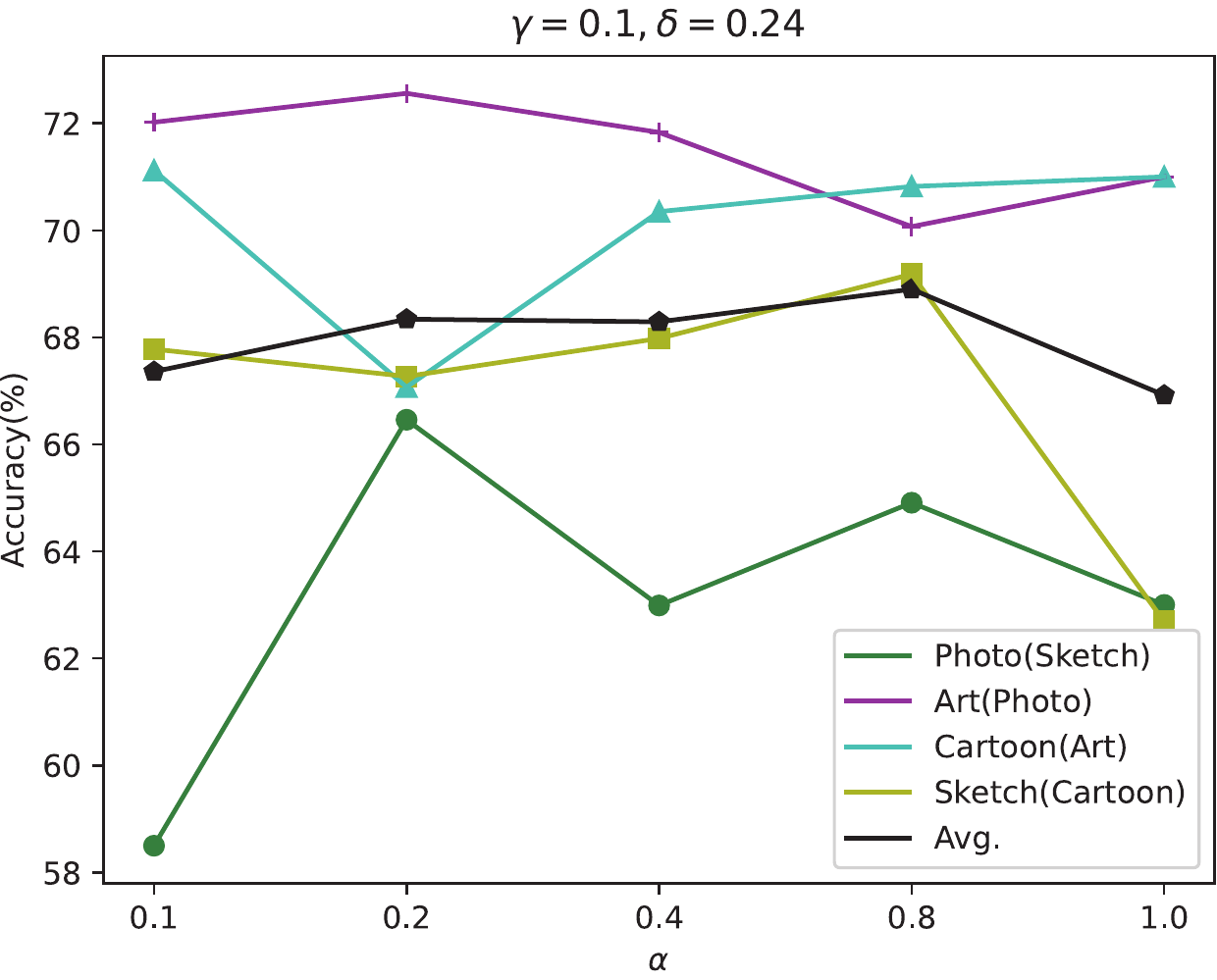}
\end{minipage}%
}
\subfigure{
\begin{minipage}[t]{0.48\linewidth}
\centering
\includegraphics[width=2.3in]{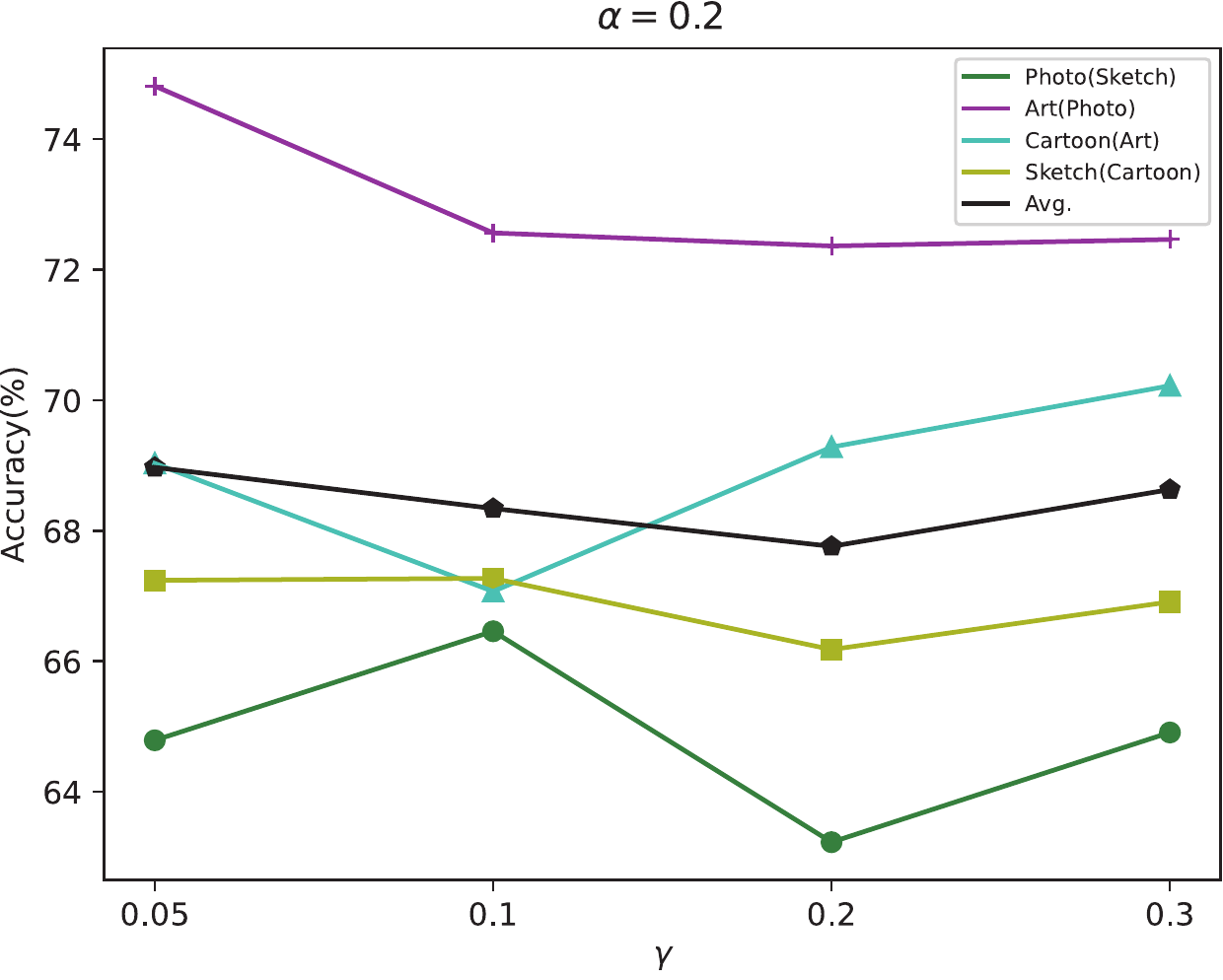}
\end{minipage}%
}%
\centering
\caption{The sensitivity analysis of $\alpha$ in mixup (left) and $\gamma$ in domain-aware pseudo-labeling (right)
on PACS.}
\label{hyper}
\end{figure}

\textbf{Sensitivity of Hyper-parameters.} 
Here we discuss the sensitivity to hyper-parameters of our method on PACS, including $\alpha$ in mixup and $(\gamma,\delta)$ in domain-aware pseudo-labeling. To simplify the analysis, We select 4 combinations. To be specific, ``Photo (Sketch)'', ``Art (Photo)'', ``Cartoon (Art)'' and ``Sketch (Cartoon)'' are chosen, where ``Photo(Sketch)'' represents the case that Sketch is labeled for training and Photo is the target. 
We train our model with $\alpha$ in [0.1, 0.2, 0.4, 0.8, 1.0]. With the increase of $\alpha$, mixup is more likely to generate more confusing samples. The results are reported in Figure \ref{hyper}. The effect of hyperparameter $\alpha$ on testing accuracy does not show similar trends in each experiment. Photo (Sketch) and Sketch (Cartoon) show large variance. And an optimal value for all experiments is between 0.2 and 0.8. 
As for $\gamma$, we set four appropriate value pairs for the weight $\gamma$ and the threshold $\delta$ in domain-aware pseudo-labeling, specifically, (0.05, 0.2), (0.1, 0.24), (0.2, 0.3) and (0.3, 0.36). It can be observed that the performance is not sensitive to $\gamma$ and smaller value is slightly better. 
We set $\alpha$ and $\gamma$ as 0.2 and 0.1 for all combinations in our experiments.

\begin{table}[tbp]
  \centering
  \caption{Compare different policies of selecting class representation in domain-aware pseudo-labeling.}
    \begin{tabular}{|c|c|c|c|c|c|}
    \hline
Target & Photo & Art & Cartoon & Sketch & Avg. \\
    \midrule
One & 83.06 & 66.57 & 61.11 & 67.95 & 69.67 \\
 Ensemble & 83.97 & 66.99 & 61.21 & 69.34 & 70.38 \\
    \hline
    \end{tabular}
  \label{policy}
  
\end{table}

\textbf{Different Schemes for Class Representation.}
We propose two policies of selecting class representation for domain-aware pseudo-labeling in our framework.
The results is shown in Table~\ref{policy}, ``One'' stands for picking up the most confident one unlabeled sample from each class as class representation at every epoch, and ``Ensemble'' means calculating the average of several samples as domain-aware class representation. 
When applying ``Ensemble'' policy, we choose one sample only if its confidence is higher than the existing highest confidence in the same class. We hold all chosen samples and calculate the average at every epoch.
This experiment shows that ``Ensemble'' policy is more fault-tolerant and its performance is better.

\begin{figure*}[htbp]
\centering
\subfigure[Labeled:Art Unlabeled:Photo,Sketch]{
\begin{minipage}[t]{0.98\linewidth}
\centering
\includegraphics[width=4in]{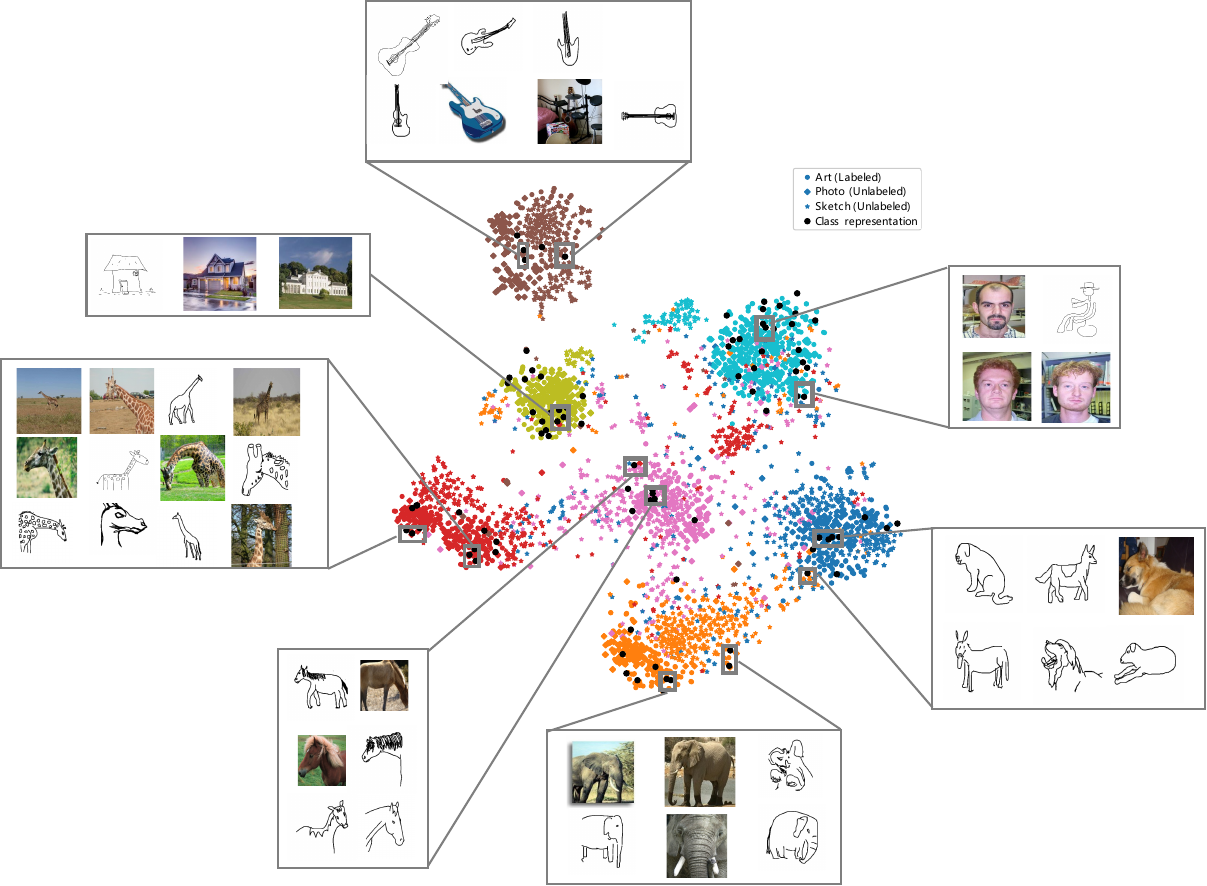}
\end{minipage}%
}\\%
\subfigure[Labeled:Cartoon Unlabeled:Photo,Art]{
\begin{minipage}[t]{0.98\linewidth}
\centering
\includegraphics[width=4in]{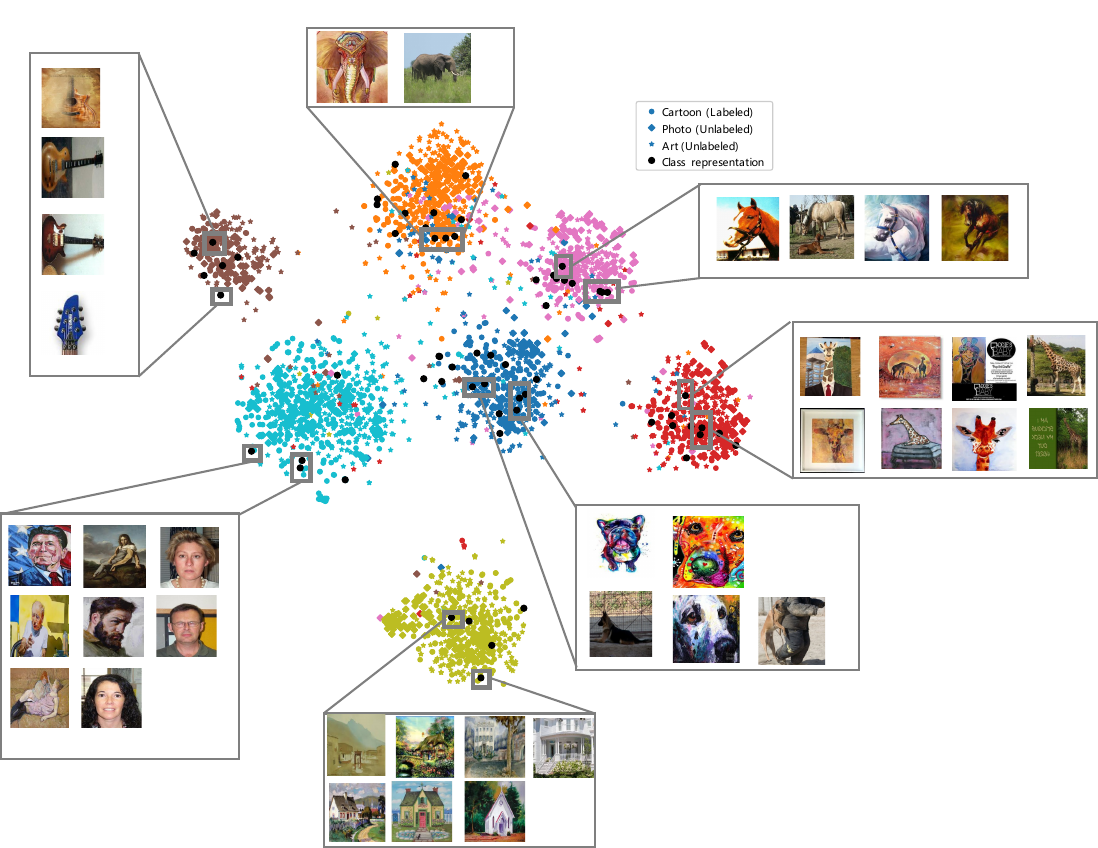}
\end{minipage}%
}%
\centering
\caption{The t-SNE \cite{tsne} visualization of feature distribution in source domains. The selected class representation of all unlabeled domains are marked and the original images are shown. Two combinations of PACS are involved.}
\label{visu}
\end{figure*}

\textbf{Visualization of Feature Distributions.}
Figure \ref{visu} visualizes the feature distributions of three source domains on PACS, including one labeled domain and two unlabeled domains. 
As seen, the selected domain-aware class representation samples are accurate according to the corresponding original images. 
Meanwhile, it can be observed that there are still some misclassified unlabeled samples, especially in Sketch, due to the large domain gap. 

\section{Conclusion}\label{s-conclusion}
In this paper, we address the problem of semi-supervised domain generalization via producing better pseudo-labels for unlabeled data. 
Firstly, we propose domain-aware pseudo-labeling for picking up more accurate pseudo labels for unlabeled data by domain-based modification. 
Then, a dual-classifier network structure is employed to promote the generalization of the model and the accuracy of pseudo-labels.
Finally, utilizing the accurate pseudo-labels in unlabeled domains, we apply domain mixup to them and enforce entropy regularization on ambiguous samples. Extensive experiments on benchmark datasets validate the efficacy of our framework.

\section*{Acknowledgement}
This work was supported by NSFC (62222604, 62192783), CAAI-Huawei MindSpore Project (CAAIXSJLJJ-2021-042A), China Postdoctoral Science Foundation Project (2021M690609), Jiangsu Natural Science Foundation Project (BK20210224), and CCF-Lenovo Bule Ocean Research Fund.





\bibliographystyle{elsarticle-num}
\bibliography{pr.bib}







\end{document}